\documentclass{article}
\usepackage{spconf,amsmath,graphicx}
\usepackage{lipsum}
\usepackage{booktabs}
\usepackage{snapshot}

\usepackage{xcolor}
\definecolor{dark_green}{rgb}{0, 0.5, 0}

\renewcommand{\eqref}[1]{Equation~\eqref{#1}}

\newcommand{\Table}[1]{Table.~\ref{tbl:#1}}

\newcommand{\Fig}[1]{Fig.~\ref{fig:#1}}

\renewcommand{\paragraph}[1]{\vspace{0.5em}\noindent\textbf{#1}}

\title{Reducing the Human Effort in Developing PET-CT Registration}
 
\name{
Teaghan O'Briain$^1$ \quad 
Kyong Hwan Jin$^2$ \quad
Hongyoon Choi$^3$ \quad 
Erika Chin$^4$}
\secondlinename{
Magdalena Bazalova-Carter$^1$ \quad
Kwang Moo Yi$^1$\thanks{This research was supported in part by NSERC USRA, NSERC Discovery Grant, and by systems supplied by Compute Canada.}
}
\address{
    $^1$University of Victoria, BC, Canada $\quad$ 
    $^2$\'Ecole Polytechnique F\'ed\'erale de Lausanne, Switzerland \\
    $^3$Seoul National University Hospital, Seoul, Korea $\quad$
    $^4$BC Cancer, BC, Canada
    }
\begin{document}
\maketitle
\begin{abstract}
    We aim to reduce the tedious nature of developing and evaluating methods for aligning PET-CT scans from multiple patient visits.
Current methods for registration rely on correspondences that are created manually by medical experts with 3D manipulation, or assisted alignments done by utilizing mutual information across CT scans that may not be consistent when transferred to the PET images.
Instead, we propose to label multiple key points across several 2D slices, which we then fit a \emph{key curve} to. 
This removes the need for creating manual alignments in 3D and makes the labelling process easier.
We use these key curves to define an error metric for the alignments that can be computed efficiently.
While our metric is non-differentiable, we further show that we can utilize it during the training of our deep model via a novel method.
Specifically, instead of relying on detailed geometric labels -- \emph{e.g.,} manual 3D alignments -- we use synthetically generated deformations of real data.
To incorporate robustness to changes that occur between visits other than geometric changes, we enforce consistency across visits in the deep network's internal representations.
We demonstrate the potential of our method via qualitative and quantitative experiments.
\end{abstract}
\begin{keywords}
    PET, CT, Image Registration, Evaluation, Weakly Supervised
\end{keywords}

\section{Introduction}
\label{sec:introduction}

Aligning PET-CT scans accurately is a crucial step in disease assessment and treatment planning.
Physicians need to be able to compare scans from one visit to another to determine how the disease has progressed.
However, in practice -- even with care -- the initial alignments are never perfect.
Therefore, semi-automatic registration methods~\cite{mim, syn} have been developed to correct this misalignment and help doctors assess the extent and progression of the disease.
More recently, there have been efforts~\cite{voxelmorph} to make use of the recent progress in deep learning for this task.
Still, in practice, we rely mostly on physicians to align different scans, as the automated methods are not accurate enough.

A potential reason behind the slow pace of development compared to other applications that use deep learning~\cite{He15},
is that creating labels in 3D is very difficult.
For example, creating ``ground truths'' for PET-CT alignments requires providing 3D correspondences.
This could be in the form of an alignment given by an expert through a graphical interface tool that manipulates data in 3D, or perhaps by a few 3D key point correspondences (again, chosen manually).
In both cases, a clinical expert needs to interact with the data in 3D, which is cumbersome and error prone as our display devices are still in 2D.
Without a substantial amount of these ground truth alignments, it becomes difficult to adopt traditional deep learning-based methods, and -- more importantly -- evaluate the alignments effectively.

To overcome the difficulty in dealing with 3D data, we propose to create annotations in 2D by interacting with slices of our 3D scans, which we then convert automatically to 3D annotations.
Specifically, we propose to annotate related keypoints across multiple 2D slices of the PET-CT scans, which we then use to fit a smooth curve, therefore creating \emph{key curves}.
We create multiple key curves for a given pair of scans and compare how well these curves are aligned.
It is worth noting that our metric does not require exact 3D correspondences, as we evaluate distances over the curves and not individual points.
For efficient and easy computation, we compute this distance numerically.

In addition to our 2D labeling strategy, we also propose a novel automated registration method based on deep learning that utilizes our metric.
As our metric is computed numerically -- thus, is non-differentiable -- we do not use it directly as our training objective and use it only to validate the accuracy of our model during  (and after) training.
Instead, we propose to train a deep network without requiring ground-truth alignments between PET-CT scans of the same patient across different visits.
In order to do this, we create random synthetic transformations on individual PET-CT scans and learn to undo the transformations as in \cite{rocco2017}, but in 3D.

Furthermore, to make our method robust to changes that occur between scans that are not due to misalignments, we enforce the intermediate representations within the deep network (the feature maps) to contain the same semantic meaning.
This is done by enforcing the linear centered kernel alignment (LCKA)~\cite{kornblith2019} -- a metric recently proposed to measure the similarity between feature maps -- to be large.
Putting this all together, our framework has the advantage that the training is purely data-driven, and human expert knowledge is only used to validate its correctness through our key curves.

We empirically verify the effectiveness of our method by first showing qualitatively how our labels and metrics relate to the actual misalignment between scans, and how the proposed method successfully reduces the alignment error. 
We further demonstrate that enforcing invariance across scans is important to achieve best performance.

\section{Related Work}

\paragraph{Metrics for evaluating PET-CT registration.}
A common procedure~\cite{fitzpatrick1998} to create ground-truth annotations for PET-CT alignment is to have clinical experts identify multiple sets of corresponding key points in different scans -- or more precisely, 2D slices of a 3D scan.
One then evaluates how accurate the registration is in terms of these key points.
However, this method is limited to points on 2D slices and cannot be used to measure 3D registration.
In addition, it is non trivial to extend to 3D; one needs to analyze the data across multiple 2D slices to identify corresponding slices, which is difficult, if not impossible.
Others~\cite{urschler2007} have proposed intensity based metrics that compare voxel-wise intensities between registered images.
This may be adequate for CT images, but because PET images do not have a ``standard'' intensity scale, these metrics cannot be easily transferred to PET images, especially in the form of a quantitative metric.

\paragraph{Methods for PET-CT registration.}
One of the main methods used in a clinical setting for PET-CT image registration is the MIM software \cite{mim}.
Image registration in MIM is done by first determining the transformation between the two CT scans using mutual information, and then applying that transformation to the PET images.
One downside of this is that it cannot focus on clinically important regions; it is based on a numerical solution which globally optimizes for all points of interest.
This can further break down when there are significant changes, especially those that are not explainable by via geometric deformations (\emph{e.g.}, the patient loses a significant amount of weight). 
Using machine learning techniques for image registration could provide a possible solution to this.

Learning-based methods have gained interest recently~\cite{lee2009learning,voxelmorph} for automating medical image registration. 
In~\cite{lee2009learning}, the authors replace traditional similarity metrics with \emph{learned} ones to increase performance. 
Their method, however, relies on the patch-based processing which is unable to capture the global structure.
In~\cite{voxelmorph}, a fully learned Convolutional Neural Network (CNN)-based framework is proposed for registering general biomedical (3D) images.
Their method is closely related to ours, but creates a deformation map that is applied locally for each voxel, whereas our method provides a global parameterized transformation.
They further do not explicitly account for robustness across patient visits.

\section{Method}

\subsection{Key-curves as geometric representations}

\begin{figure}
  \centering
    \includegraphics[width=0.9\linewidth, height=0.6\linewidth]{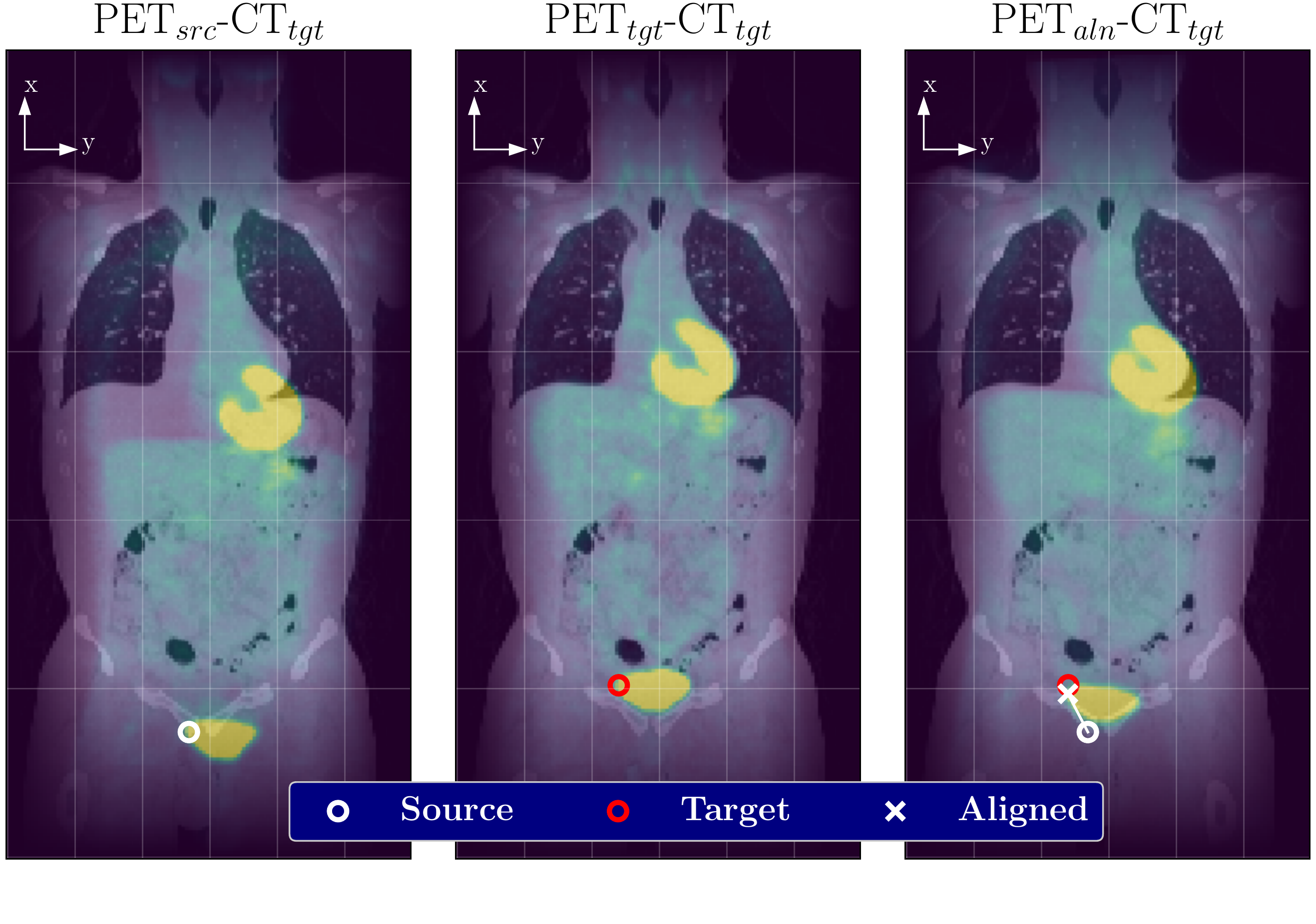}
    \vspace{-1em}
    \caption{An example of the selection of a key point for different visits of the same patient.
    To visualize how good/bad one scan is aligned with another, we show PET scans of one scan (\emph{src}) overlayed on top of CT scans of another visit (\emph{tgt}).
    Note that on the left --  when simply overlaying PET scans with CT scans from different visits without alignment -- the bone locations do not correspond properly with the bladder; the one in the center shows properly aligned PET and CT scans. 
    These could be aligned well with our method as shown on the right.
    }
    \vspace{-0.5em}
    \label{fig:key_point}
\end{figure}

To overcome the difficulties in dealing with 3D data, we propose to use a representation that is not subject to these difficulties: a 3D curve.
Given enough corresponding 3D curves between two visits from the same patient, it is possible to measure the misalignment between the two scans, just by measuring the distance between these curves.
In more detail, we represent curves in 3D as a second order polynomial that outputs $x$ and $y$ coordinates given a $z$ coordinate.
We then compute the distance between the two curves by computing the distance for selected $z$ slices and averaging them.

To obtain key curves we rely on 2D annotations. 
As shown in the example in \Fig{key_point} (left and center), one needs to only label matching points in 2D.
Note that the two scans do not have to correspond to the exact same $z$ slice since with curves -- unless the curve is perfectly straight, which in practice does not happen -- the average distance will only be zero if and only if the scans are perfectly aligned.
To facilitate easy labeling, we have further developed a simple interactive software with a graphical user interface.

\begin{figure}
  \centering
    \includegraphics[width=0.9\linewidth,height=0.4\linewidth]{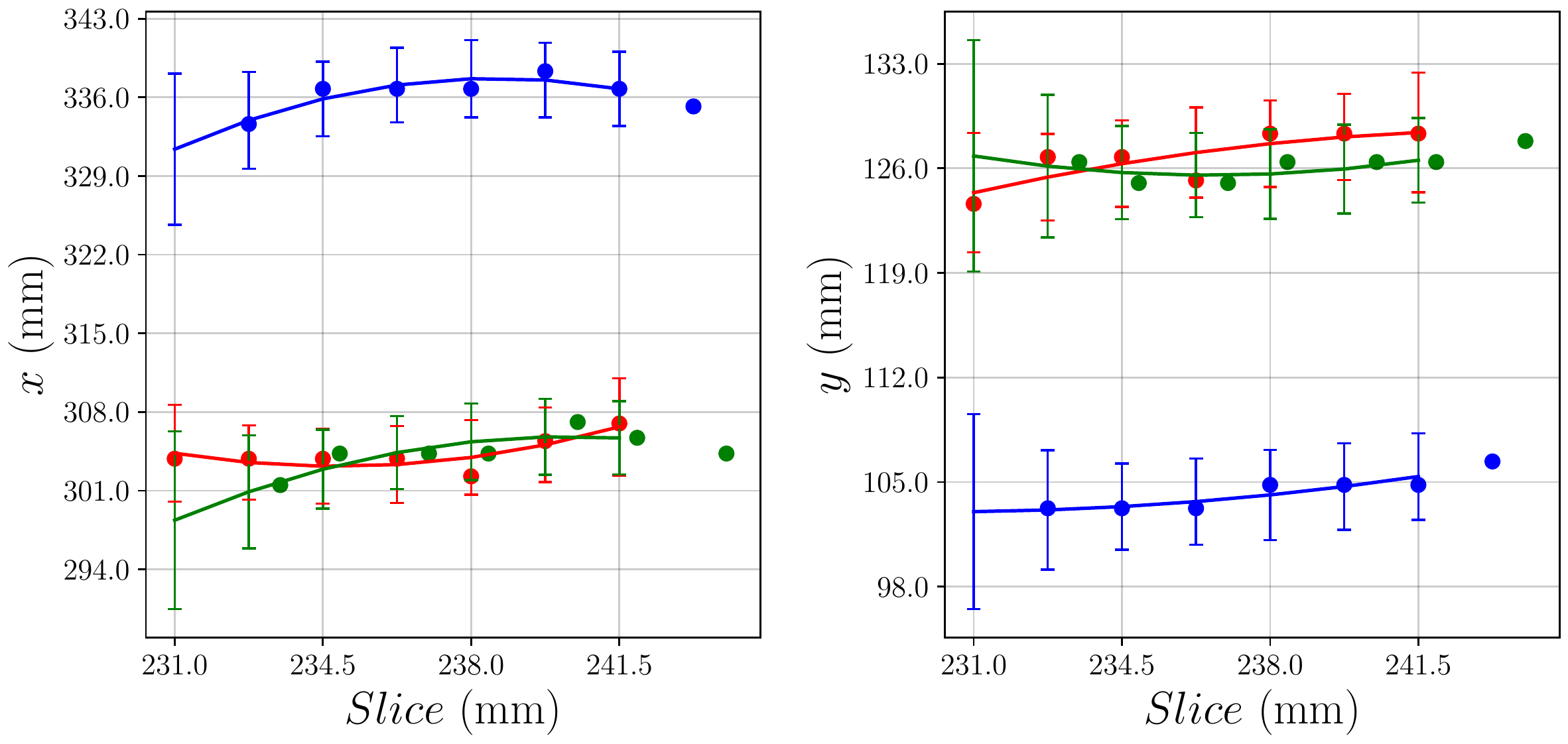}
    \vspace{-0.5em}
    \caption{
    An example of annotated key points and the fit curves.
    For easy visualization, we show the curve and points projected to $x$ and $y$ axes.
    We further show the uncertainty at each point as error bars.
    We show results for two different scans which we label \textcolor{blue}{source} and \textcolor{red}{target}, as well as the source scan \textcolor{dark_green}{aligned} to the target scan.
    }
    \label{fig:key_feature}
    \vspace{-0.5em}
\end{figure}

\paragraph{Incorporating uncertainty.} 
All labeling methods inherently create uncertainty, which we incorporate for visual inspection, as shown in \Fig{key_feature}.
We estimated the uncertainty in the key point selection by assessing the variance in our ability to annotate individual points.
The average \emph{selection uncertainty} for our dataset is 2.52 mm and 1.96 mm in the $x$ and $y$ direction, respectively.

Another source of uncertainty comes from the fact that not all $z$ slices have key points to select (for example, the bladder does not extend to the back bone).
As in~\cite{wolberg2006}, we model the confidence on the curves to be poorer in these regions where we extrapolate, which can be seen in \Fig{key_feature}.

\subsection{Novel method for PET-CT image registration}

As discussed previously, the 
key curve metric itself is convenient for evaluating alignments, however, it is non-differentiable.
We therefore propose a novel weakly supervised method that allows easy utilization of the proposed metric.

In detail, instead of depending on geometric labels, we rely on synthetic transformations of scans (i.e. self-supervision) as in~\cite{rocco2017}, but extend the method to 3D from the original 2D setup.
After applying synthetic transformations, we then train a deep network to recover the transformation parameters by just looking at the scans before and after the transformation.
We use both PET and CT scans as the input to our framework by concatenating the two as a multi-channel 3D image.
The dynamic range of PET scans can differ greatly from one scan to another, and we therefore pre-process our scans by computing the gradient of the log of the PET images.

In order to predict the transformations, we first apply 9 consecutive 3D ResNet blocks~\cite{He16} to the pre-processed scans to \emph{extract} features from both the ``source'' and ``target'' (transformed) scans.
We then create a similarity matrix, as in \cite{rocco2017}, by flattening the feature maps of both our source and target scans, then performing a cross product between the two.
This similarity matrix is then provided as an input to a successive 3D CNN that regresses to the transformation parameters.
As in \cite{rocco2017}, we first apply an affine transformation for the initial alignment, then a thin plate spline (TPS) transformation, which accounts for the remaining -- more subtle -- misalignments.
We also found it important to to train these two steps separately, as our experiments with both being trained simultaneously did not give promising results.

However, the self-supervised approach alone does not provide robustness, as each training sample consists only of a single PET-CT scan along with the transformed version of that image.
We therefore enforce consistency among the feature representations (created by the deep network) across different PET-CT scans of the same patient.
In more detail, we enforce this on the feature maps -- used to create the similarity matrix -- by maximizing the recently proposed linear centered kernel alignment (LCKA)~\cite{kornblith2019}.
The LCKA encodes the similarity of deep feature responses, while ignoring their geometrical configuration.
Lastly, by evaluating our key curve metric on a validation set throughout training, we selectively stop training when the performance starts to decline (i.e. when we over-fit).

\section{Results}

\begin{figure}
  \centering
    \includegraphics[width=0.9\linewidth, trim = 0 0 0 65, clip]{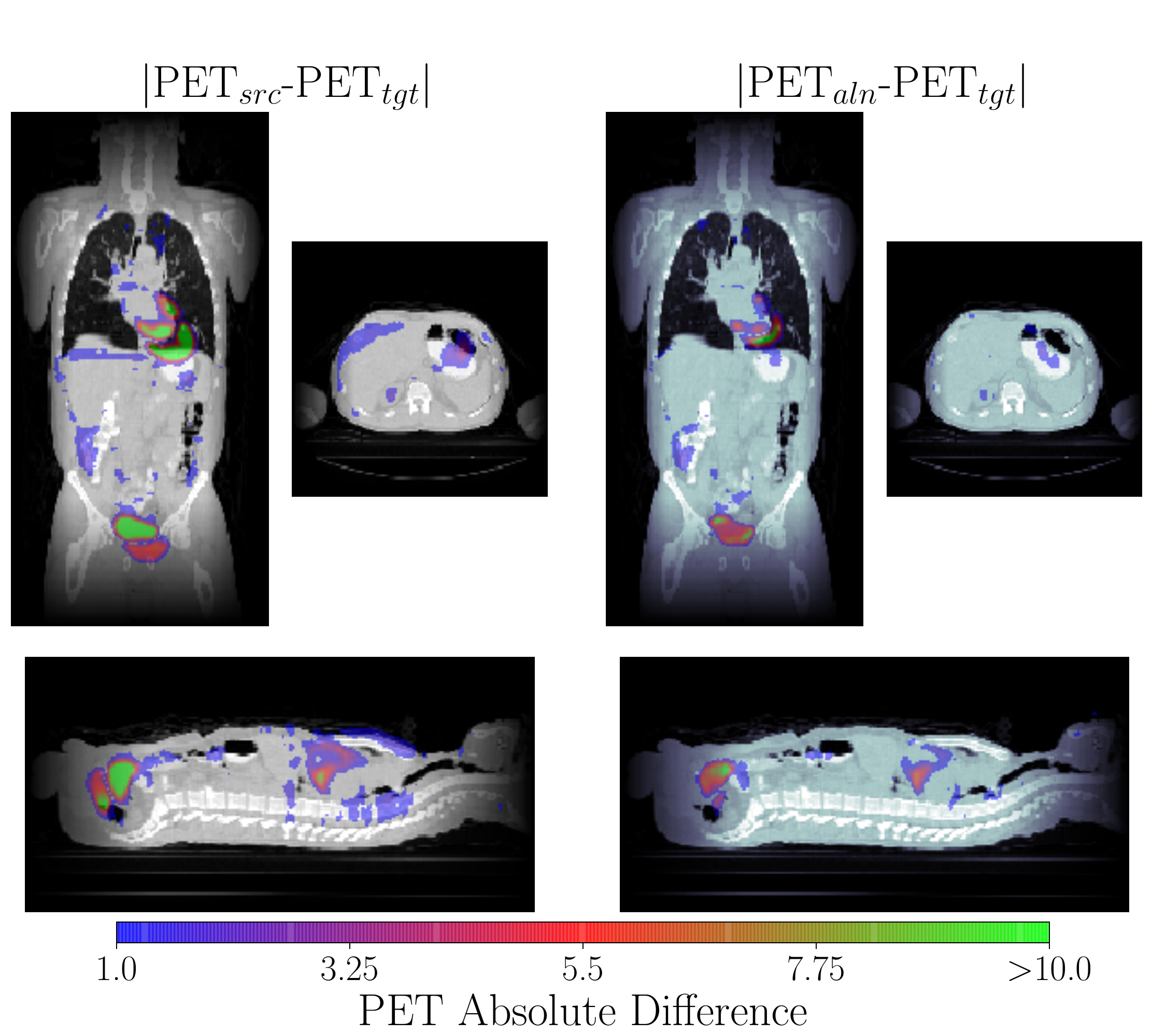}
    \vspace{-0.5em}
    \caption{
    Example residual image showing the difference between PET scans when they are (left) not aligned and (right) aligned with our method.
    Notice how the misalignment near the spine  and the ribs has disappeared in the aligned image.
    }
    \label{fig:qualitative}
    \vspace{-0.5em}
\end{figure}

We apply our method to the QIN-HEADNECK dataset \cite{PETdata, Fedorov2016, Clark2013} of PET-CT images, which consists of 651 studies on 156 patients. 
We selected 296 patient scans that are of reasonable size (larger than 120$\times$120$\times$240 pixels at a resolution of 3.5mm).
We further split this sample into training (267), validation (15), and test (14) sets based on the patient ID.
We emphasize here that we only label the validation and test pairs with our key curves, and for training, we strictly rely \emph{only} on the pairwise relationship and no geometrical supervision.

\subsection{Qualitative results}

In \Fig{key_point} we show an example of slices from aligned and unaligned scans, where we annotate key locations within these slices. 
Without any registration, the scans can initially be quite misaligned, despite the efforts that were taken to have the patient lie in the same position.
After applying our registration method, visual comparison shows that the alignment drastically improves.
\Fig{key_feature} further shows the alignment between two scans in terms of a particular key curve. 
Finally, we show a residual image comparison in \Fig{qualitative}. 

\begin{table}
    \caption{Quantitative results. 
    Root Means Squared Error (RMSE) of 22 key curves selected in the test set of 14 images (10 pairs).
    Our method performs best.
    }
    \label{tbl:quantitative}
    \begin{center}
    \begin{tabular}{ c c c c }
    \toprule 
         & Unaligned  & w/o LCKA    & Our Method \\
    \midrule 
    RMSE (mm)       & 29.25 & 15.89 & {\bf 13.41} \\
    \bottomrule
    \end{tabular}
    \end{center}
    \vspace{-1em}
\end{table}

\subsection{Quantitative results}
In \Table{quantitative}, we apply the key curve metric to a variety of different visit pairs in our test set, and report the summary of how the proposed registration method performs.
It is important to note that without LCKA -- which would become a 3D version of \cite{rocco2017} -- the method is not able to learn to be invariant across visits, and therefore provides worse results. 
Our method
provides improved results.
\section{Conclusion}

We have proposed a novel annotation scheme for labeling geometrical 
correspondences in PET-CT scans based on key curves.
The method allows labeling to be performed easily in 2D, and relieves the need for manipulating data in 3D, which can be tedious.
Through our weakly-supervised registration method, we show that we can train a deep model with this metric, even though the metric is non-differentiable.
Furthermore, we show empirically how the metric and the annotations can be used, as well as the effectiveness of our method.

A limitation of our method is that the key curve-based metric is non-differentiable. 
However, an Expectation Maximization (EM) style optimization can be applied to circumvent this.
This is part of our immediate future work, and we expect it to increase the usefulness of our annotation method, also allowing easy creation of geometrically labeled data that can be used to further enhance our registration method.

\bibliographystyle{IEEEbib}
\bibliography{string,references,kwang_paper,biomed,vision}

\end{document}